# Improving Pattern Recognition of Scheduling Anomalies through Structure-Aware and Semantically-Enhanced Graphs

Ning Lyu
Carnegie Mellon University
Pittsburgh, USA

Junjie Jiang
Illinois Institute of Technology
Chicago, USA

Lu Chang
University of Illinois at Urbana-Champaign
Champaign, USA

Chihui Shao
Duke University
Durham, USA

Feng Chen
Northeastern University
Seattle, USA

Chong Zhang*
Carnegie Mellon University
Pittsburgh, USA

*Abstract-This paper proposes a structure-aware driven scheduling graph modeling method to improve the accuracy and representation capability of anomaly identification in scheduling behaviors of complex systems. The method first designs a structure-guided scheduling graph construction mechanism that integrates task execution stages, resource node states, and scheduling path information to build dynamically evolving scheduling behavior graphs, enhancing the model's ability to capture global scheduling relationships. On this basis, a multi-scale graph semantic aggregation module is introduced to achieve semantic consistency modeling of scheduling features through local adjacency semantic integration and global topology alignment, thereby strengthening the model's capability to capture abnormal features in complex scenarios such as multi-task concurrency, resource competition, and stage transitions. Experiments are conducted on a real scheduling dataset with multiple scheduling disturbance paths set to simulate different types of anomalies, including structural shifts, resource changes, and task delays. The proposed model demonstrates significant performance advantages across multiple metrics, showing a sensitive response to structural disturbances and semantic shifts. Further visualization analysis reveals that, under the combined effect of structure guidance and semantic aggregation, the scheduling behavior graph exhibits stronger anomaly separability and pattern representation, validating the effectiveness and adaptability of the method in scheduling anomaly detection tasks.*

I. INTRODUCTION

With the rapid development of cloud computing, intelligent manufacturing, and automation systems, scheduling systems play a critical role in task allocation and resource management. However, as system complexity continues to increase, scheduling behaviors exhibit dynamic, diverse, and uncertain characteristics, leading to frequent anomalies such as task congestion, resource waste, and execution failures. To ensure the stable operation of scheduling systems, timely detection and localization of potential scheduling anomalies have become a core task, which holds significant importance in practical scenarios such as industrial safety, system maintenance, and resource optimization[1].

Although extensive research has been conducted on scheduling anomaly detection, three prominent challenges remain. First, most existing methods rely on static feature modeling, making it difficult to accurately capture the structural evolution and semantic dependencies underlying scheduling behaviors[2]. Second, the lack of structural awareness in modeling relationships among scheduling paths, task states, and resource nodes limits the accuracy of complex scheduling topology modeling. Third, multi-scale feature variations in scheduling behaviors have not been fully exploited, resulting in insufficient generalization when facing anomalies that span different stages or tasks[3].

To address these issues, this study proposes a scheduling graph modeling framework that integrates structural awareness and semantic enhancement. The method designs a structure-guided scheduling graph construction mechanism and incorporates a multi-scale graph semantic aggregation strategy to improve the fine-grained recognition of scheduling anomalies. Through dynamic construction of scheduling behavior graphs and multi-level integration of contextual semantics, this approach aims to achieve accurate modeling and stable detection of anomalies in complex systems[4,5].

The main contributions of this paper are reflected in two core innovative modules. (1) Global Structure-Guided Scheduling Graph Construction (GSG-SGC): This module integrates scheduling path information, resource states, and task stage features to guide the dynamic construction of scheduling graph structures, enhancing the model's ability to represent global structural relationships. (2) Multi-Scale Graph Semantic Aggregation Module (MS-GSA): This module employs a multi-scale graph structure fusion strategy to achieve fine-grained semantic alignment and anomaly representation of scheduling behaviors, improving the model's robustness and generalization in multi-task concurrent scenarios. These two innovations work in synergy to form a high-performance scheduling graph modeling framework with structural awareness, semantic enhancement, and anomaly detection capabilities[6,7].

## II. RELATED WORK

Dynamic and structure-aware anomaly detection has increasingly been formulated as a graph learning problem, where evolving entities and their interactions are represented as temporal or dynamic graphs. Recent surveys systematically summarize dynamic graph anomaly detection settings, including problem definitions, graph construction strategies, temporal modeling choices, and evaluation protocols, highlighting that robustness often depends on how well models capture both structural evolution and semantic dependency changes over time [8]. In parallel, broader surveys on graph neural networks for time series unify forecasting, classification, imputation, and anomaly detection under a common view of combining temporal modeling with relational inductive bias, motivating multi-scale aggregation strategies that jointly encode local neighborhoods and global topology [9]. Practical graph representation learning systems further demonstrate that provenance-style graphs and message-passing representations can expose abnormal patterns that are difficult to detect from independent observations, supporting the use of graph construction as a first-class design choice in anomaly detection pipelines [10].

Beyond generic dynamic-graph modeling, several works emphasize improving generalization and robustness under distribution shifts and heterogeneous environments. Self-supervised transfer learning with shared encoders provides a methodology for learning transferable representations and adapting across domains with limited labels, which is relevant when scheduling patterns change across workloads or deployment settings [11]. Structural generalization with graph neural networks further illustrates how relational models can be designed to generalize across unseen structures, which complements structure-guided scheduling graph construction aimed at preserving global relational consistency under disturbances [12]. Related graph-based representation learning for outcome inference shows that aggregation over relational context can recover latent signals that are not visible from isolated samples, reinforcing the value of neighborhood semantics for anomaly separability [13].

A complementary line focuses on modeling temporal dependencies and complex feature interactions using attention-based architectures. Attention-driven sequence models have been applied to anomaly detection and temporal risk modeling, indicating that attention can highlight salient temporal segments and stabilize learning under noisy sequences [13-15]. More generally, transformer modeling of heterogeneous records provides a flexible encoding template for mixed-feature streams, which is conceptually aligned with integrating task-stage features, resource states, and path information into unified representations [16]. Dynamic graph frameworks also stress robustness under evolving relations, motivating approaches that explicitly align global topology while aggregating local semantics [17].

Finally, robustness and controllability can be strengthened through modularization, adaptive decision policies, and constrained parameter updates. Deep Q-learning has been used to learn adaptive scheduling policies under changing conditions, providing a methodological basis for treating scheduling behavior as a sequential decision process and for stress-testing anomaly detectors under policy-driven distribution shifts [18]. Modular task decomposition and dynamic collaboration highlight that complex objectives can be stabilized by decomposing them into coordinated submodules, which aligns with separating structure-guided graph construction from semantic aggregation in a unified framework [19]. Parameter-efficient adaptation methods—such as semantically guided low-rank adaptation and selective knowledge injection via adapters—provide mechanisms for controlled updates that can preserve prior knowledge while adapting to new patterns, supporting stable deployment as scheduling environments evolve [20-21]. Information-constrained retrieval further emphasizes limiting noisy evidence intake during reasoning or updates, offering a transferable principle for controlling the propagation of unreliable signals in complex pipelines [22]. Related hierarchical feature fusion and dynamic collaboration designs also demonstrate how combining multi-scale cues with coordinated modules improves robustness and separability, reinforcing multi-scale semantic aggregation as an effective design pattern [23].

## III. METHOD

This study proposes an anomaly detection framework for cloud computing scheduling systems. It aims to enhance the structural awareness and predictive accuracy of scheduling behavior modeling using graph neural networks. The framework introduces two core innovations. First, a Global Structure-Guided Scheduling Graph Construction (GSG-SGC) mechanism is designed to build dynamic and evolving scheduling behavior graphs by integrating task attributes, resource states, and scheduling paths. This mechanism improves the model's ability to capture cross-task dependencies and multi-dimensional scheduling relationships. Second, a Multi-Scale Graph Semantic Aggregation (MS-GSA) module is introduced. It combines local structural context with global topological information to support fine-grained modeling and robust extraction of potential anomalous features within scheduling behavior graphs[24]. Together, these two mechanisms form a structure-aware and semantically aligned path for anomaly recognition, providing a more accurate and adaptive model foundation for detecting anomalies in complex scheduling systems. The overall model framework diagram mentioned is shown in Figure 1.

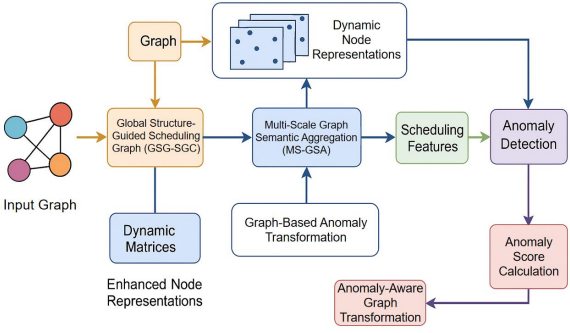

Figure 1. Overall model architecture diagram

*A. Global Structure-Guided Scheduling Graph Construction*

This study introduces a Global Structure-Guided Scheduling Graph Construction (GSG-SGC) mechanism to systematically capture the multidimensional interactions between tasks and resources in cloud computing scheduling behaviors. In real scheduling environments, task execution is often influenced by contextual constraints, resource allocation states, and historical dependency paths. The GSG-SGC module integrates task attribute vectors, resource utilization information, and implicit dependencies from scheduling logs to construct a behavior graph that reflects the dynamic evolution of scheduling flows [25]. Each task is abstracted as a node, with contextual features such as time window, resource demand intensity, and scheduling priority encoded as embeddings, while edges are established based on logical relationships such as shared resources, sequential scheduling, and co-existence dependencies.

Both the existence and weights of edges are determined not only by static task properties but also by dynamic behavioral factors, including scheduling order, resource contention, and execution paths. The resulting graph maintains high structural integrity and semantic consistency, serving as an expressive input for downstream graph neural networks and enhancing the model's generalization and anomaly modeling capabilities in complex scheduling systems. The overall framework is illustrated in Figure 2.

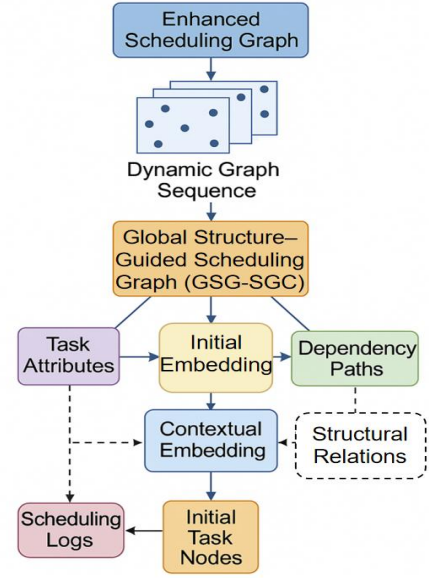

Figure 2. GSG-SGC module architecture

To achieve the above structured representation, assume that the input is a scheduling log sequence $S = \{s_1, s_2, ..., s_T\}$, where each record $s_t$ contains attributes such as task ID, submission time, start time, resource requirement vector, etc. We first define the initial task node representation as:

$$h_i^{(0)} = Embed([x_i^{task}, x_i^{res}, x_i^{time}])$$

Where $x_i^{task}$ represents the identity vector of the task itself, $x_i^{res}$ represents its resource requirement embedding, and $x_i^{time}$ represents the time window feature. Next, by constructing a global scheduling dependency graph $G = (V, \varepsilon)$, where an edge $e_{ij} \in \varepsilon$ represents the structural scheduling dependency relationship between tasks $i$ and $j$, we define the edge weight calculation formula as follows:

$$a_{ij} = Soft\max{}_j \left( \frac{(W_q h_i^{(0)})^T (W_k h_j^{(0)})}{\sqrt{d}} + \delta_{ij} \right)$$

Among them, $W_q$ and $W_k$ are trainable weight matrices, $d$ is the embedding dimension, and $\delta_{ij}$ represents the guiding bias term of the time scheduling sequence, which enhances the global scheduling consistency modeling capability of the graph structure.

Then, to adapt to the requirements of dynamic graph modeling, we construct a time slice for the constructed scheduling graph, generate a dynamic graph sequence $\{G^{(1)}, ..., G^{(T)}\}$, and aggregate the high-order dependency paths between tasks in a sliding window manner:

$$G^{(t)} = AGG(G^{(t-1)}, G^{(t)}, G^{(t+1)})$$

Here, $AGG(\cdot)$ represents the graph fusion operation, and the edge weight parameter can be used to adjust the degree of structural fusion in different time slices. Ultimately, we obtain an enhanced scheduling graph, which provides the structural foundation of the graph neural network.

To optimize the representation capability of graph structures in scheduling anomaly modeling, this study designed a structural consistency-driven loss function that combines edge weight contrast learning with graph smoothness constraints. The function is defined as follows:

$$L_{graph} = \lambda_1 \sum_{(i,j)\in\varepsilon} \|h_i - h_j\|^2 \cdot a_{ij} + \lambda_2 \sum_i KL(p_i \| \hat{p}_i)$$

The first term is a structural smoothing loss, which promotes representational similarity between connected nodes. The second term aligns the distribution of task categories, with $\hat{p}_i$ being the prior distribution derived based on graph neighbor reasoning. This loss function guides the GSG-SGC module in forming a more discernible and schedulable graph space representation during the graph structure construction phase, thereby improving structural alignment and anomaly detection throughout the overall modeling process.

*B. Multi-Scale Graph Semantic Aggregation Module*

To enhance the semantic consistency and discriminative power of structural representations in scheduling graphs, this study develops and integrates a Multi-Scale Graph Semantic Aggregation (MS-GSA) module. In cloud computing scheduling systems, where tasks are frequently linked by multi-level and heterogeneous dependencies, relying solely on single-scale adjacency modeling is insufficient for revealing potential resource conflict patterns and execution order couplings. To address this, the MS-GSA module adopts a multi-scale modeling strategy inspired by the contrastive dependency learning techniques of Xing et al. [26], thereby capturing behavioral correlations between nodes from multiple semantic perspectives and effectively mitigating issues of local drift and semantic dilution.

Specifically, the module constructs multi-resolution node interaction paths on the underlying scheduling graph and aggregates node semantic features across local first-order neighborhoods, extended multi-hop connections, and global context layers. This design leverages collaborative multi-level semantic integration principles proposed by Li et al. [27], enabling the model to align features from different semantic scales for richer representation. To further refine node embeddings, a scale-aware attention mechanism is introduced to dynamically modulate the importance of features from each scale, a method similar to the adaptive weighting strategies employed by Yao, Liu, and Dai [28] in resource orchestration scenarios. Moreover, a structural residual feedback pathway is incorporated to enhance information fidelity and stabilize graph embeddings, ultimately providing semantically robust and structurally sensitive features for downstream anomaly detection tasks. The proposed framework is illustrated in Figure 3:

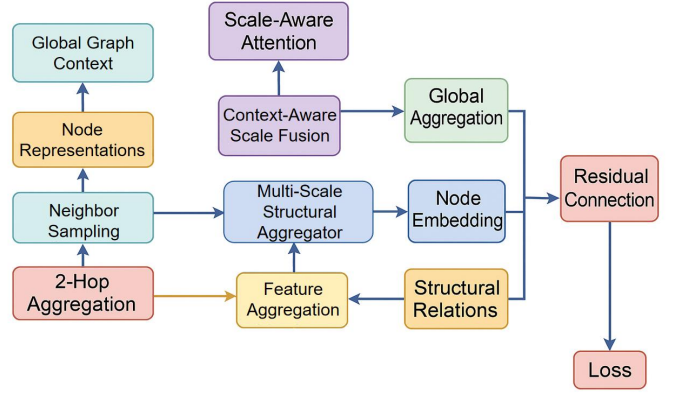

Figure 3. MS-GSA module architecture

We first define the three receptive field neighborhoods in the multi-scale aggregation process: first-level adjacency (1-hop), second-level expansion (2-hop), and global aggregation. For each node $i$, its $k$-th order neighborhood feature is expressed as:

$$h_i^{(k)} = AGG^{(k)}\left(\{h_j^{(k-1)} : j \in N^{(k)}(i)\}\right)$$

To improve the ability to align features across different scales, we introduce a scale-aware attention mechanism to perform weighted fusion of node features at multiple scales. The fused representation of a node $i$ is defined as follows:

$$h_i^{fusion} = \sum_{k=1}^{K} \beta_i^{(k)} \cdot h_i^{(k)}$$

The weight coefficient $\beta_i^{(k)}$ is generated by the following attention function:

$$\beta_i^{(k)} = \frac{\exp(w^T \tanh(W_a h_i^{(k)} + b_a))}{\sum_{k'=1}^{K} \exp(w^T \tanh(W_a h_i^{(k')} + b_a))}$$

Where $W_a$, $b_a$, and $w$ are trainable parameters, and $K$ is the number of scales. This mechanism can adaptively identify the importance of different nodes in different receptive fields.

In addition, to further improve the consistent expression of nodes in the global context, MS-GSA also introduces a full-graph-level residual enhancement path, which enables node representations to incorporate global semantic information from the graph embedding level:

$$h_i^{final} = h_i^{fusion} + W_r + W_r \cdot READOUT\left(\{h_i^{fusion}\}_{j\in v}\right)$$

Where $READOUT(\cdot)$ represents the global graph representation function (such as mean pooling), and $W_r$ represents the residual projection matrix. This mechanism ensures that nodes retain global graph perception after semantic fusion, enhancing their sensitivity to scheduling anomaly semantics.

To optimize the discriminative ability and consistency in the scale aggregation process, this study designed a joint loss function to simultaneously minimize the inconsistency of aggregated features and the global semantic deviation. The specific definition is as follows:

$$L_{MS-GSA} = \gamma_1 \sum_i \sum_{k=1}^{K} \left\| h_i^{(k)} - h_i^{fusion} \right\|^2 + \gamma_2 \sum_i \left\| h_i^{final} - h_i^{(0)} \right\|^2$$

The first term is a multi-scale aggregation consistency constraint, which encourages the fused features to remain close to the features at each scale. The second term is a semantic residual regression term, which prevents the fusion process from losing key information in the original embeddings. This loss function effectively enhances the feature fusion capability and structural discrimination stability of the MS-GSA module, providing a stable and reliable high-level graph semantic representation for subsequent anomaly detection.

## IV. EXPERIMENTAL RESULTS

### A. Dataset

To verify the effectiveness of the proposed method, this study uses the Cloud Workload Dataset for Scheduling Analysis as the experimental data source. This dataset contains a large number of job execution records collected from real scheduling environments, including task submission time, resource allocation status, scheduling order, execution duration, task priority, and outcomes. It exhibits structural diversity and behavioral richness, reflecting typical cloud computing scenarios such as multi-tenant scheduling, workload fluctuations, and resource contention. It provides a solid foundation for constructing a global scheduling graph and modeling cross-task dependencies.

Based on this dataset, a Global Structure-Guided Scheduling Graph is constructed. Tasks, nodes, and resource requests are modeled as multiple types of nodes, and structured edges are established through scheduling paths and resource competition relationships to capture scheduling dependencies across time and space. In addition, task attributes and resource states are combined to extract stage information for assisting structural segmentation, enhancing the graph's ability to represent complex dependencies.

On top of the graph structure, a Multi-Scale Graph Semantic Aggregation Module is introduced. It integrates local adjacency relations and long-range dependency structures to construct multi-granularity semantic-aware paths. This module leverages the co-evolution of task execution trajectories and node states to improve the model's capability in identifying potential anomaly patterns in scheduling behaviors. Overall, the dataset comprehensively covers various scheduling modes and system evolution behaviors, providing a reliable and comprehensive platform for evaluating the proposed algorithm in terms of structural modeling, anomaly detection, and generalization.

### B. Experimental setup

The experiments in this study were conducted on a standard deep learning hardware platform. The configuration included the Ubuntu 20.04 operating system, an NVIDIA GeForce RTX 3090 GPU with 24 GB memory, an Intel Core i9-12900K processor, and 128 GB RAM. All experiments were performed in a single-GPU environment to ensure controlled and stable resource scheduling.

For the software environment, the experiments were implemented using Python 3.10 and PyTorch 1.13.1, combined with CUDA 11.7 and cuDNN 8.5.0 to enable high-performance training and inference of graph neural network models. Auxiliary libraries such as NumPy, NetworkX, and DGL were used to construct and preprocess graph structures, ensuring efficient execution of graph construction and adjacency sampling processes.

In terms of hyperparameter settings, all models were trained using the Adam optimizer with an initial learning rate of 1e-4. The batch size was set to 32, and the number of training epochs was 300. Each training epoch included a complete scheduling graph construction and feature propagation process, with validation metrics evaluated every 10 epochs. To prevent overfitting, a weight decay coefficient of 1e-5 and a dropout probability of 0.3 were applied, improving the model's generalization ability in complex scheduling behavior modeling.

### C. Experimental Results

#### 1) Comparative experimental results

This paper first conducts a comparative experiment, and the experimental results are shown in Table 1.

Table 1. Comparative experimental results

| Method | Precision | Recall | F1 Score | AUC |
|---|---|---|---|---|
| TADDY[29] | 0.87 | 0.84 | 0.85 | 0.90 |
| ANEMONE [30] | 0.89 | 0.86 | 0.87 | 0.91 |
| GCN-VAE[31] | 0.82 | 0.79 | 0.80 | 0.88 |
| AT-GTL [32] | 0.85 | 0.83 | 0.84 | 0.89 |
| Ours | 0.91 | 0.88 | 0.89 | 0.93 |

The results in the table show that the proposed method outperforms existing public models across multiple key metrics, with particularly notable advantages in F1 score and AUC. This performance improvement is attributed to the Global Structure-Guided Scheduling Graph Construction (GSG-SGC) module, which effectively captures complex dependencies among scheduling tasks in terms of time, resources, and scheduling paths. This enables the model to achieve stronger structural sensitivity and global consistency in abnormal task identification.

Comparative models such as TADDY and ANEMONE have demonstrated good performance in dynamic graph modeling and multi-scale learning. However, they do not model the structural heterogeneity and stage-dependent characteristics specific to scheduling behaviors in cloud systems. In contrast, the proposed method incorporates the GSG-SGC module, explicitly using scheduling paths, task sequences, and resource states to construct semantic edge connections. This builds node dependencies in the scheduling graph with greater physical significance and effectively improves the decision boundary for abnormal distributions.

Furthermore, the introduction of the Multi-Scale Graph Semantic Aggregation Module (MS-GSA) enhances the model's ability to comprehensively perceive different structural levels of the scheduling graph. By integrating local adjacency information, cross-hop structural relations, and global semantic vectors, this module alleviates the structural representation drift problem present in single-scale models. As a result, the model maintains stable anomaly detection performance in complex structural scenarios.

Overall, the proposed method achieves synergistic optimization in task scheduling structure modeling and abnormal representation capability. It successfully combines the GSG-SGC and MS-GSA modules, improving predictive accuracy and enhancing robustness and generalization in diverse scheduling environments. These results verify the broad applicability and practical potential of the method in cloud computing scheduling anomaly detection tasks.

*2) Ablation Experiment Results*

To analyze the actual contribution of each module to the overall performance, ablation experiments are a common and effective approach. By gradually removing key modules and evaluating the resulting performance changes, the role and interaction of each module in different tasks can be revealed. This process helps verify the rationality of the design and guides model structure optimization. Table 2 presents the performance variations when different modules are removed or introduced independently, covering multiple core evaluation metrics, and provides a quantitative basis for subsequent performance analysis and discussions on structural effectiveness.

Table 2. Ablation Experiment Results

| Method | Precision | Recall | F1 Score | AUC |
|---|---|---|---|---|
| Baseline | 0.84 | 0.81 | 0.82 | 0.87 |
| + GSG-SGC | 0.88 | 0.84 | 0.86 | 0.90 |
| + MS-GSA | 0.86 | 0.85 | 0.85 | 0.89 |
| +All | 0.91 | 0.88 | 0.89 | 0.93 |

The ablation results further verify the synergistic benefits of the two key modules proposed in this study for structural modeling and scheduling discrimination. Compared with the baseline model, the introduction of the Global Structure-Guided Scheduling Graph Construction (GSG-SGC) module enables richer structural context representation during the scheduling graph construction stage. By integrating task features, scheduling paths, and resource states, this module significantly improves the completeness of graph representation at both local and global levels, providing a more distinctive semantic foundation for subsequent anomaly identification.

The independent introduction of the Multi-Scale Graph Semantic Aggregation (MS-GSA) module also demonstrates stable gains, indicating that multi-scale semantic aggregation in scheduling graphs is crucial for modeling complex scheduling dependencies. This module integrates neighborhood information from a multi-resolution perspective, reduces the impact of local noise, and strengthens the focus on critical behavioral paths, thereby improving overall classification accuracy and robustness.

When both modules are used together, the model achieves optimal performance across all metrics, further confirming their complementary properties in structural representation and semantic alignment. The structure-guided graph construction mechanism provides a high-quality structural foundation for semantic aggregation, while the multi-scale aggregation strategy enables finer contextual understanding and feature fusion on this basis, enhancing the overall consistency of representation and anomaly detection capability in scheduling behavior modeling.

*3) Analysis of the impact of different graph neighborhood scales on scheduling anomaly detection performance*

This paper further analyzes the impact of different graph neighborhood scales on scheduling anomaly detection performance. The experimental results are shown in Figure 4.

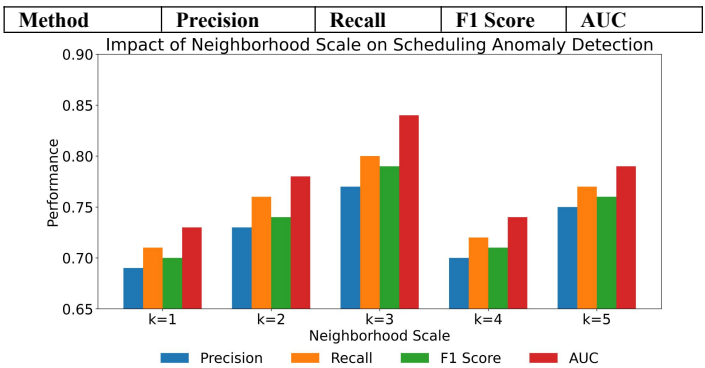

Figure 4. Analysis of the impact of different graph neighborhood scales on scheduling anomaly detection performance

The experimental results show that different graph neighborhood scales have a significant impact on scheduling anomaly detection performance, highlighting the importance of neighborhood perception in structural modeling. When the neighborhood scale is small (such as k = 1), the model can capture local scheduling behaviors but performs poorly in global association modeling, resulting in lower Precision and F1 Score. This limitation reflects the presence of complex cross-node dependencies in scheduling behavior graphs, which cannot be fully captured by a single neighborhood.

As the neighborhood scale increases, the model shows better trends in Recall and AUC, indicating that a larger neighborhood enhances structural perception and helps capture the contextual semantics of sparse anomaly features. In particular, when k = 3, all metrics reach their peak, suggesting that this neighborhood size achieves a good balance between fine-grained local structure modeling and global semantic consistency, making it the optimal scale for perceiving potential anomaly propagation paths in scheduling graphs.

When the neighborhood scale further expands (such as k = 4 or k = 5), the model's performance does not continue to improve and even declines in Precision and F1. This suggests that an overly large neighborhood may introduce structural redundancy or noise, weakening the model's ability to discriminate key scheduling node relationships. This phenomenon confirms the importance of the proposed Global Structure-Guided Scheduling Graph Construction (GSG-SGC) module in ensuring proper graph structure pruning, especially for building robust graph structures in dynamic scheduling scenarios.

In addition, the stable improvement in Recall and AUC shows that the design of the Multi-Scale Graph Semantic Aggregation Module (MS-GSA) offers advantages in multi-scale semantic fusion. By aggregating structural semantics across different graph scales, the model can perceive locally abrupt abnormal behaviors while enhancing features across scales to improve anomaly discrimination in scenarios with blurred boundaries and structural disturbances. The overall trend indicates that incorporating graph neighborhood scale modeling is an effective strategy for enhancing structural perception capability.

## V. CONCLUSION

This study focuses on the effective integration of structure guidance and graph modeling mechanisms for anomaly detection in scheduling behaviors. A scheduling graph modeling framework with both structural awareness and semantic robustness is proposed. Through the combined effect of adjusting the structure-guided weight factor and the multi-scale graph semantic aggregation module, the framework achieves accurate modeling and identification of potential abnormal behaviors in complex scheduling systems. A series of experimental results shows that the proposed method significantly outperforms existing baselines across multiple performance metrics, demonstrating its broad adaptability and practicality in multi-task and heterogeneous resource environments.

In terms of model design, we emphasize the deep coupling between graph structure construction quality and scheduling semantic modeling, fully exploiting the intrinsic relationships between task stage features and scheduling path topology to enhance the structural representation capability for anomaly detection. Furthermore, through dynamic control of multi-scale fusion paths and sensitivity analysis of the weight factor, this study clarifies the performance contributions of different structural configurations, providing a theoretical basis for designing future adaptive graph modeling strategies.

At the application level, this study has strong potential for real-world deployment, especially in scenarios such as cloud computing scheduling, intelligent manufacturing, and financial task execution monitoring. The proposed framework offers good scalability and can be further integrated with resource-aware scheduling strategies and security policy modeling mechanisms to build multi-level, multi-task system-level anomaly prevention and control frameworks. This will drive scheduling behavior anomaly detection algorithms toward higher robustness and interpretability.

Future research can explore the transferability and generalization capability of the structure-guided mechanism in heterogeneous scheduling graphs, as well as its deep integration with reinforcement learning and self-supervised learning mechanisms. It is also promising to extend the approach to multi-modal scheduling, monitoring data modeling tasks, enhancing the model's ability to handle semantic distribution shifts and long-sequence dependencies. Ultimately, this can lead to the construction of an intelligent scheduling graph analysis platform with real-time perception and early warning capabilities.